\documentclass[lettersize,journal]{IEEEtran}
\usepackage{amsmath,amsfonts}
\usepackage{algorithmic}
\usepackage{array}
\usepackage[caption=false,font=normalsize,labelfont=sf,textfont=sf]{subfig}
\usepackage{textcomp}
\usepackage{stfloats}
\usepackage{url}
\usepackage{verbatim}
\usepackage{graphicx}
\usepackage{xcolor}

\usepackage{hyperref}
\usepackage{multirow}
\usepackage{lettrine}

\usepackage{booktabs}
\usepackage{caption}
\def\BibTeX{{\rm B\kern-.05em{\sc i\kern-.025em b}\kern-.08em
    T\kern-.1667em\lower.7ex\hbox{E}\kern-.125emX}}
\usepackage{balance}
\begin{document}
\title{MambaRain: Multi-Scale Mamba-Attention Framework for 0-3 Hour Precipitation Nowcasting}

\author{
\IEEEauthorblockN {
   Chunlei Shi$^{1,2}$,
   Cui Wu$^{3}$,
   Xiang Xu$^{4}$,
   Hao Li$^{2}$,
   Ni Fan$^{5}$,
   Xue Han$^{1}$,
   Yongchao Feng$^{6}$,
   Yufeng Zhu$^{1}$,
   Boyu Liu$^{1}$,
   Zengliang Zang$^{7}$,
   Hongbin Wang$^{8}$,
   Yanlan Yang$^{1}$
   Dan Niu$^{1,\dagger}$
}
\thanks{This work was supported by the  Heavy Rainfall Research Foundation of China (No. BYKJ2025M14), China Meteorological Administration Xiong\textrm{'}an Atmospheric Boundary Layer Key Laboratory (No. 2025LABL-B12), and by the National Natural Science Foundation of China (62374031, 62331009), and by NSFC-Jiangsu Province (BK20240173).}

\thanks{
$^{\dagger}$ Corresponding author(email:230238514@seu.edu.cn).\\
$^{1}$School of Automation, Southeast University;
$^{2}$Nanjing XinDa Institute of Meteorological Science and Technology;
$^{3}$Beijing Leninainfo Technology Co., Ltd.;
$^{4}$China CEC Engineering Corporation;
$^{5}$School of Mathematical Sciences, Tongji University;
$^{6}$State Key Laboratory of Virtual Reality Technology and Systems, Beihang University;
$^{7}$College of Meteorology and Oceanography, National University of Defense Technology;
$^{8}$Key Laboratory of Transportation Meteorology of China Meteorological
Administration, Nanjing Innovation Institute for Atmospheric Sciences;
}
}

\markboth{IEEE GEOSCIENCE AND REMOTE SENSING LETTERS,~Vol.~x, No.~x, September~202x}%
{MambaRain: Multi-Scale Mamba-Attention Framework for 0-3 Hour Precipitation Nowcasting}

\maketitle

\begin{abstract}
Accurate precipitation nowcasting over extended horizons (0-3 hours) is crucial for disaster prevention and real-time decision-making, yet remains a critical challenge. 
Existing deterministic methods typically focus on shorter periods (0-1 or 0-2 hours) and exhibit rapid performance degradation beyond 90 minutes due to difficulties in capturing long-range spatiotemporal dependencies from radar observations.
To address these limitations, we propose MambaRain, a novel multi-scale encoder-decoder framework that integrates Mamba's linear-complexity long-range temporal modeling with self-attention mechanisms.
Our core innovation is a hybrid architecture where Mamba blocks capture global temporal evolution across extended sequences via selective state space mechanisms, while self-attention complements this by explicitly modeling spatial correlations within precipitation fields—a capability absent in Mamba's sequential processing. This synergy enables comprehensive spatiotemporal modeling, extending the effective forecasting window to 2-3 hours with improved accuracy.
Furthermore, we introduce a spectral loss to mitigate blurring effects in chaotic precipitation systems, preserving fine-scale motion details.
Unlike previous deterministic approaches that struggle with long-range spatiotemporal dependencies, MambaRain achieves superior performance in 0-3 hour nowcasting, with significant accuracy gains especially in the 2-3 hour range.
Extensive experiments on Xinjiang and Southeast China SWAN datasets demonstrate substantial improvements over existing methods in both precision and efficiency. Our project is available at \url{https://spring-lovely.github.io/MambaRain2025/}.
\end{abstract}

 
\section{Introduction}
\label{intro}

\lettrine[lines=2]{P}{recipitation} nowcasting—the generation of 
high-resolution rainfall intensity forecasts for the upcoming 0--3 
hours—plays an indispensable role in weather-dependent 
decision-making across aviation safety, agricultural planning, flood 
control, and urban infrastructure management~\cite{shi2026wavec2r, lin2025alphapre, gao2025lmcast, yu2025pimmnet, allen2025end, niu2025m4caster,chang2026future}. In China, the increasing 
frequency and intensity of convective precipitation events driven by 
climate change have amplified the demand for reliable extended-horizon 
forecasts, particularly in topographically complex regions such as 
Xinjiang and Southeast China where orographic effects strongly modulate 
rainfall patterns. Despite decades of progress in both numerical weather 
prediction and data-driven approaches, extending the effective 
forecasting window beyond 90 minutes while preserving spatial fidelity 
and computational efficiency remains a fundamental and largely unresolved 
challenge in operational meteorology~\cite{Cui2025Pre, Zhang2026How, Qian2025Dete, Pan2025Joint, Wang2025prer}.

Traditional operational nowcasting systems, including optical 
flow-based methods (e.g., OptFlow)~\cite{zhang2020learnable} and variational extrapolation 
approaches (e.g., Sprog)~\cite{reinoso2022nationwide}, estimate radar-derived 
precipitation motion fields and advect them forward in time. While 
computationally efficient, these methods operate under a linear motion 
assumption that renders them inherently incapable of representing 
non-linear phenomena such as convective initiation, dissipation, and 
cell merging—processes that increasingly dominate precipitation evolution 
beyond the first 30--60 minutes. The fundamental physical barrier is 
that precipitation systems are governed by chaotic atmospheric dynamics, 
where small perturbations in thermodynamic instability or moisture 
convergence lead to rapidly diverging future states, making purely 
deterministic extrapolation unreliable at extended lead times.

Deep learning approaches have emerged as powerful alternatives by 
learning complex spatiotemporal patterns directly from large radar 
observation archives. ConvRNN-based models, including 
ConvLSTM~\cite{shi2015conv} and TrajGRU~\cite{shi2017shi}, introduced 
recurrent architectures for sequential radar echo processing, while 
CNN-Transformer hybrids such as AA-TransUnet~\cite{yang2022aa} and 
Earthformer~\cite{gao2022earthformer} leveraged attention mechanisms 
for richer spatiotemporal feature extraction. More recently, 
NowcastNet~\cite{zhang2023now} integrated physics-constrained 
generation for improved structural realism. However, a common failure 
mode persists across these deterministic approaches: optimization with 
mean squared error (MSE) loss in the spatial domain causes models to 
converge toward the \emph{conditional mean} of the inherently 
multi-modal precipitation distribution, producing over-smoothed, blurry 
predictions at longer lead times where forecast uncertainty is high. 
This regression-to-the-mean effect becomes increasingly severe beyond 
the 90-minute mark, rendering forecasts meteorologically unrealistic 
and operationally misleading—particularly for high-intensity convective 
events that matter most for hazard warnings.

To mitigate the blurring problem, generative approaches have been 
applied to precipitation nowcasting. Diffusion-based models such as 
DiffCast~\cite{yu2024diffcast}, CasCast~\cite{gong2024cascast}, and 
MTLDM~\cite{chaorong2024extreme}, along with GAN-based methods like 
FsrGAN~\cite{niu2024fsrgan}, have demonstrated improved perceptual 
quality and probabilistic forecast skill. More recent innovations 
include simulation-driven convective uncertainty 
modeling~\cite{yin2025simcast}, large-model-inspired multi-scale 
sequence fusion~\cite{gao2025lmcast}, and multi-modal radar-satellite 
dual-stream frameworks~\cite{xu2025syncast}. Despite these advances, 
generative models typically require tens to hundreds of iterative 
denoising or generation steps, resulting in inference latencies on the 
order of several seconds (e.g., DiffCast requires $\sim$6.3s per 
sample)—far exceeding the sub-second requirements of real-time 
operational systems. Moreover, achieving accurate and temporally 
consistent predictions at the 2--3 hour horizon remains challenging 
even for these sophisticated approaches.

A key but frequently overlooked factor in extended-horizon nowcasting 
is the influence of underlying terrain on precipitation dynamics~\cite{zheng2026pre, Jian2026Radio, Luo2026Pre, Chen2020Resolve}. 
Topographic features such as mountain ranges, valleys, and coastlines 
create persistent orographic lifting, rain shadow effects, and 
thermally driven local circulations that fundamentally shape where 
and how precipitation develops and intensifies. In topographically 
complex regions such as Xinjiang, which is bordered by the Tianshan 
and Kunlun mountain ranges, and Southeast China, characterized by 
the Nanling Mountains and irregular coastlines, ignoring terrain 
information introduces systematic spatial biases that purely temporal 
or spectral improvements cannot fully compensate. Incorporating 
Digital Elevation Model (DEM) data as an auxiliary geographic input 
provides the model with static orographic context that anchors 
precipitation predictions to the underlying landscape, yet this 
approach remains surprisingly underexplored in the existing 
nowcasting literature.

Meanwhile, the emergence of state space models (SSMs), particularly 
the Mamba architecture~\cite{gu2023mamba}, has opened new possibilities 
for efficient long-range sequence modeling. Unlike Transformers whose 
self-attention mechanism scales quadratically with sequence length 
($\mathcal{O}(L^2)$), Mamba employs a selective state space mechanism 
that achieves linear complexity ($\mathcal{O}(L)$) while maintaining 
competitive long-range memory. This property is highly desirable for 
modeling the temporal evolution of precipitation systems across 0--3 
hour sequences comprising dozens of radar frames~\cite{Zhao2025Pre, Jin2026Feature, Li2025ADNM-UNet, Wu2025WeatherGen, He2026Hi-RSMamba}. However, Mamba's 
inherently sequential scan-based processing is better suited for 
capturing temporal dependencies than for modeling the two-dimensional 
spatial correlations within individual radar frames. This observation 
motivates a hybrid design that leverages Mamba for efficient global 
temporal memory while retaining self-attention for parallel spatial 
reasoning, thereby combining the complementary strengths of both 
paradigms.

To address these co-existing challenges, we propose 
\textbf{MambaRain} (see Fig.~\ref{figov}), a novel multi-scale 
encoder-decoder framework for accurate and efficient 0--3 hour 
precipitation nowcasting. MambaRain synergistically integrates 
Mamba's linear-complexity long-range temporal modeling with 
self-attention's parallel spatial reasoning, enhanced by terrain-aware 
DEM encoding and a spectral loss function that targets the root cause 
of prediction blurring. Our framework specifically addresses three 
co-existing limitations of current methods: (1)~the quadratic 
computational cost of Transformer-based temporal modeling on long 
sequences, (2)~the absence of explicit spatial dependency modeling 
in Mamba's sequential processing, and (3)~the spatial-domain MSE 
blurring effect—tackling each with a dedicated, principled design 
component. The major contributions of this paper are summarized as 
follows:

\begin{figure*}[ht]  
\centering  
\includegraphics[width=0.9\textwidth]{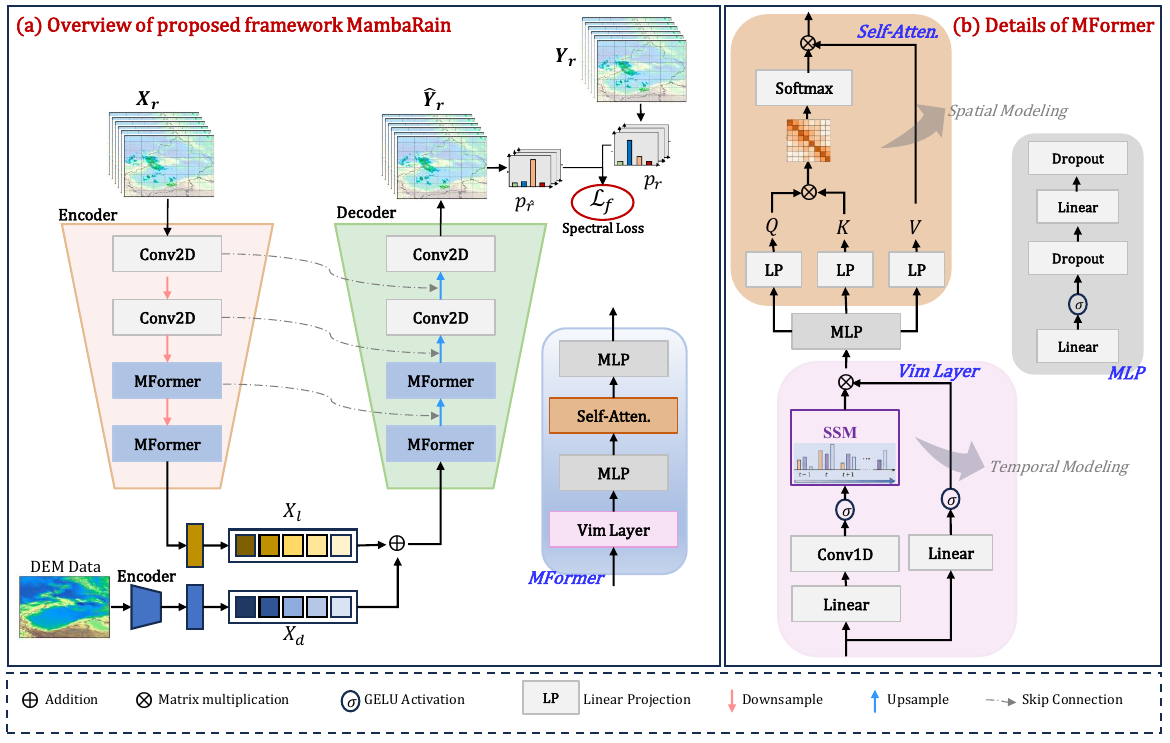}   
\caption{(a) Overview of the MambaRain model architecture featuring multi-scale encoder-decoder structure with MFormer blocks and DEM integration for terrain-aware precipitation nowcasting. (b) Details of MFormer Block showing the hybrid Mamba-Transformer architecture that combines temporal memory capabilities with spatial attention mechanisms.}
\label{figov}  
\vskip -0.23in
\end{figure*}

\begin{itemize}
    \item We propose \textbf{MambaRain}, a novel multi-scale 
    Mamba-Attention encoder-decoder framework that integrates Mamba's 
    efficient long-range temporal modeling with self-attention mechanisms 
    and DEM geographic encoding, extending reliable precipitation 
    nowcasting to the challenging 2--3 hour horizon with competitive 
    computational efficiency.

    \item We introduce \textbf{MFormer}, a hybrid Mamba-Transformer 
    block that resolves the complementarity between Mamba's sequential 
    global temporal memory and self-attention's parallel spatial 
    dependency modeling, enabling comprehensive multi-scale 
    spatiotemporal representation learning without incurring quadratic 
    complexity overhead.

    \item We design a \textbf{spectral loss function} operating in the 
    Fourier frequency domain to explicitly penalize discrepancies in 
    high-frequency precipitation structures, mitigating the spatial 
    averaging blur effect while preserving fine-scale convective 
    patterns and sharp precipitation boundaries across extended 
    forecast horizons.

\end{itemize}

\section{Method}
\subsection{Formulation of Precipitation Nowcasting}
Given a sequence of historical radar observations $\mathbf{X} = \{\mathbf{x}_1, \mathbf{x}_2, \ldots, \mathbf{x}_T\}$ spanning the past 0-1 hour, our goal is to predict future radar reflectivity fields $\hat{\mathbf{Y}} = \{\hat{\mathbf{y}}_{T+1}, \hat{\mathbf{y}}_{T+2}, \ldots, \hat{\mathbf{y}}_{T+K}\}$ for the subsequent 0-3 hour horizon that approximate the ground truth radar sequences $\mathbf{Y} = \{\mathbf{y}_{T+1}, \mathbf{y}_{T+2}, \ldots, \mathbf{y}_{T+K}\}$. The extended precipitation nowcasting task is formulated as:
\begin{equation}
    \hat{\mathbf{Y}} = f_{\theta}(\mathbf{X}), \quad \theta^* = \arg\min_{\theta} \mathbb{E} \big[ \mathcal{L}(f_{\theta}(\mathbf{X}), \mathbf{Y}) \big],
\end{equation}
where $f_{\theta}$ is a learnable radar extrapolation function that maps historical observations to future precipitation fields, $\mathcal{L}$ is a task-specific loss function, and $K$ represents the number of future time steps corresponding to the 0-3 hour forecasting horizon.

\subsection{Framework of MambaRain}
Fig. 1(a) illustrates the overall architecture of MambaRain, which comprises an encoder-decoder structure with integrated DEM feature encoding for enhanced precipitation nowcasting through multi-scale spatiotemporal modeling. 

The encoder consists of four layers: the first two layers use convolutional blocks and downsampling for low-level feature extraction, while the latter two layers incorporate MFormer modules that integrate Mamba blocks for capturing global spatiotemporal dependencies across extended sequences with self-attention mechanisms for complementary parallel spatial reasoning. The decoder maintains symmetric structure with upsampling operations and MFormer modules in the corresponding layers. Skip connections fuse multi-scale encoder outputs with decoder inputs to preserve low-level features.

DEM geographic information is encoded and integrated at the bottleneck layer for terrain-aware precipitation modeling. The framework employs a spectral loss function to mitigate blur effects and preserve fine-scale motion patterns over the 0-3 hour forecasting horizon.

\subsection{MFormer Details}
The MFormer block represents the core innovation of our framework (see Fig. 1(b)), designed as a hybrid architecture that synergistically integrates Mamba's long-range temporal memory capabilities with self-attention mechanisms for comprehensive spatiotemporal feature extraction.

\subsubsection{Feature Integration and Processing Pipeline}
At the encoder's final layer, radar input features are processed to generate output features $X_l$ with dimensions $L \times D$, where $L$ represents the sequence length and $D$ denotes the feature dimension. Simultaneously, DEM data undergoes encoding through a dedicated encoder to extract topographic information, followed by tokenization to produce $X_d$ with compatible dimensions. The integrated features $X_l + X_d$ serve as input to the decoder's final MFormer layer.

The decoder's output is computed as:
\begin{align}
    \mathbf{X}_{out} &= \text{Up}(\text{MFormer}(X_l + X_d)),
\end{align}
where the MFormer processing follows a sequential pipeline: MambaVision layer, MLP, self-attention, and final MLP layer, before upsampling.

\subsubsection{Selective State Space Mechanism}
The MambaVision component employs a selective state space model (SSM) to capture global spatiotemporal dependencies across extended sequences efficiently. The SSM computation is formulated as:
\begin{equation}
\begin{aligned}
h_k &= \bar{A} h_{k-1} + \bar{B} x_k, \\
y_k &= C h_k + D x_k, \\
\bar{A} &= e^{A \Delta}, \\
\bar{B} &= (e^{A \Delta} - I) A^{-1} B \approx \Delta B,
\end{aligned}
\end{equation}
where $B, C \in \mathbb{R}^{D \times N}$ act as projection matrices, $h_k$ represents the hidden state at step $k$, and $\Delta$ is the discretization parameter. This enables efficient processing of extended sequences with linear computational complexity, providing the temporal memory capabilities essential for extended horizon forecasting.

\subsubsection{Self-Attention Mechanism}
Following the MambaVision processing and an intermediate MLP layer, features undergo self-attention computation to enable explicit spatial dependency modeling and parallel global context aggregation. The self-attention mechanism is formulated as:
\begin{align}
    \mathcal{A} &= \text{softmax} \left( \frac{\mathbf{Q}\mathbf{K}^T}{\sqrt{d}} \right) \mathbf{V}_l,
\end{align}
where $\mathbf{Q} = \mathbf{X}\mathbf{W}_Q$, $\mathbf{K} = \mathbf{X}\mathbf{W}_K$, $\mathbf{V}_l = \mathbf{X}\mathbf{W}_{V}$, and $\mathbf{W}_Q$, $\mathbf{W}_K$, $\mathbf{W}_{V}$ are learnable weight matrices. The complete MFormer operation is expressed as:
\begin{equation}
\text{MFormer}(X_l + X_d) = \text{MLP}(\mathcal{A}(\text{MLP}(\text{ViM}(X_l + X_d)))),
\end{equation}
where ViM denotes the MambaVision layer. This sequential processing ensures that Mamba's sequential temporal processing is complemented by self-attention's parallel spatial reasoning capabilities, while integrating terrain-aware geographic information with precipitation dynamics.

\subsection{Spectral Loss}
To address the averaging blur effect in extended precipitation nowcasting, we introduce a spectral loss function that operates in the Fourier frequency domain. Traditional MSE loss optimizes for pixel-wise accuracy, inherently favoring smooth predictions over realistic precipitation boundaries with sharp transitions.

Precipitation systems exhibit frequency-dependent characteristics: high-frequency components encode sharp boundaries and convective structures, while low-frequency components represent large-scale motions. Recent studies~\cite{zhao2024mdtnet} have demonstrated that frequency-domain optimization can effectively mitigate spatial averaging effects by preserving spectral characteristics across different scales. We formulate the spectral loss as:
\begin{equation}
\mathcal{L}_{\text{spectral}} = \frac{1}{N} \sum_{n=1}^{N} \left\| \mathcal{F}(\hat{\mathbf{y}}_n) - \mathcal{F}(\mathbf{y}_n) \right\|_2^2
\label{eq:spectral_loss}
\end{equation}
where $\mathcal{F}(\cdot)$ denotes the 2D FFT, $\hat{\mathbf{y}}_n$ and $\mathbf{y}_n$ are predicted and ground truth fields, and $N$ is the number of samples. This frequency-domain optimization preserves fine-scale motion patterns and maintains global trend consistency over the 0-3 hour horizon.
\section{Experiments}
\subsection{Implementation Details}
\begin{figure}[ht]
\centering  
\includegraphics[width=0.5\textwidth]{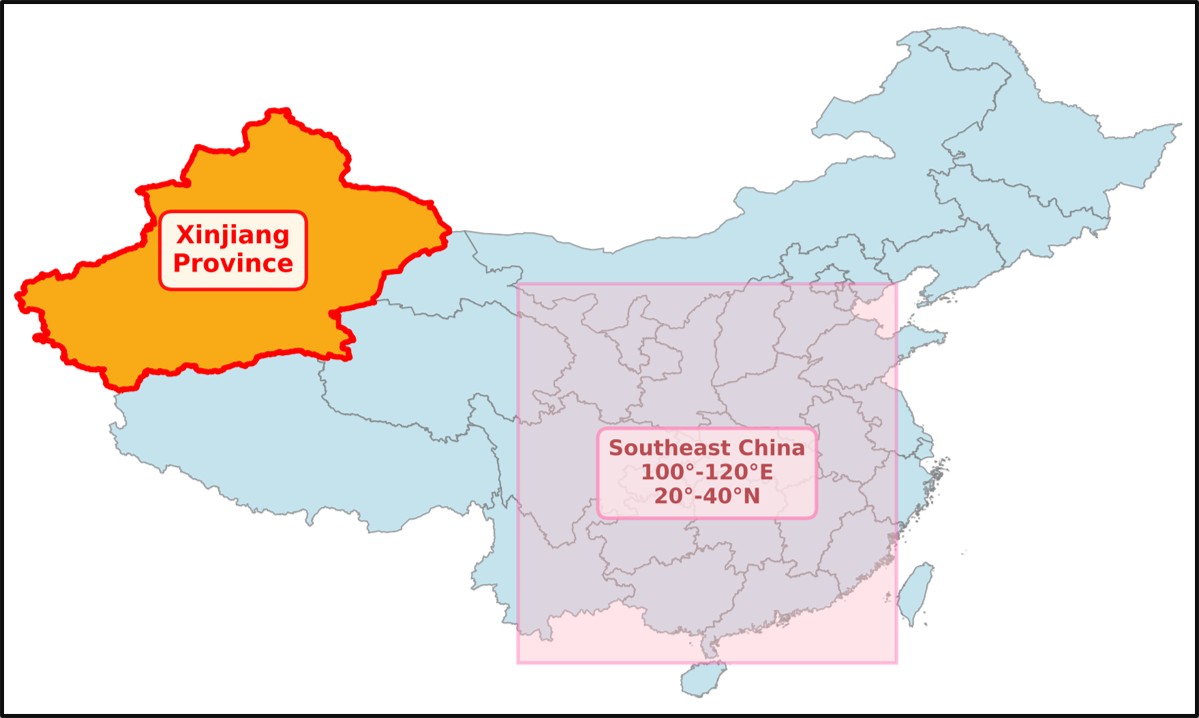}  
\caption{Study Area}
\label{fig:map}
\vspace{-2.5mm}
\end{figure}

\begin{figure}[ht]
\centering  
\includegraphics[width=0.5\textwidth]{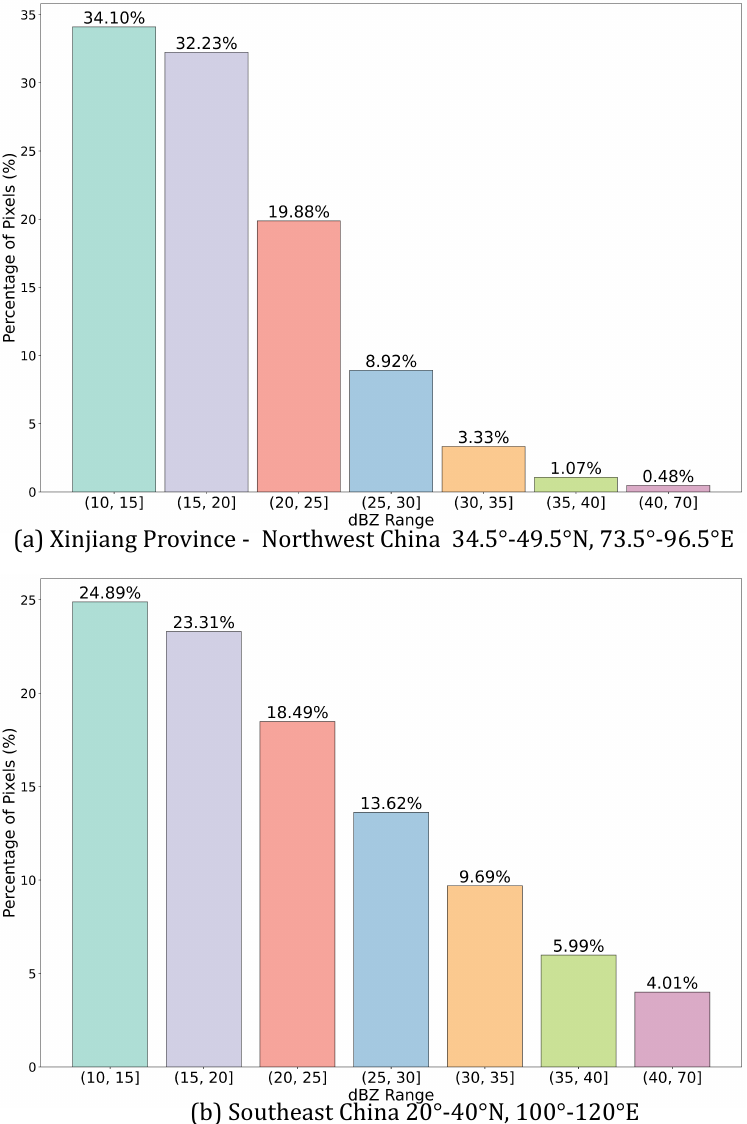} 
\caption{Echo intensity distribution above 10dBZ for both datasets.}
\label{fig:data_anly}
\vspace{-3mm}
\end{figure}

\noindent\textbf{Dataset.} We utilize the SWAN radar dataset from two geographically diverse regions: Xinjiang (73.5°-96.5°E, 34.5°-49.5°N) with mountainous terrain and Southeast China (100°-120°E, 20°-40°N) with varied topography. Both datasets have 0.04° spatial resolution with dimensions of 375×575 and 500×500 respectively (see Fig.~\ref{fig:map}). Fig.~\ref{fig:data_anly} presents the precipitation intensity distribution across both datasets.

\noindent\textbf{Training Setup.} MambaRain is trained for 250 epochs with batch size 16 on four NVIDIA RTX 3090 GPUs. The Adam optimizer is used with learning rate 0.0001 and cosine annealing scheduler for stable convergence.

\noindent\textbf{Evaluation Metrics.} 
To ensure fair comparison with existing methods, we adopt 
the same evaluation protocol as FACL~\cite{yan2024fourier}. 
Meteorological forecast skill is assessed using the Critical 
Success Index (CSI), Equitable Threat Score (ETS), False 
Alarm Ratio (FAR), and Probability of Detection (POD) at 
multiple reflectivity thresholds ($\gamma \geq$ 10, 15, 20, 
and 30\,dBZ). These threshold-based metrics are defined using 
the contingency table elements (hits $H$, misses $M$, false 
alarms $F$):
\begin{equation}
    \text{CSI} = \frac{H}{H + M + F}, \quad
    \text{FAR} = \frac{F}{H + F}, \quad
    \text{POD} = \frac{H}{H + M}.
\end{equation}
For image quality assessment, we employ the Structural 
Similarity Index (SSIM), Peak Signal-to-Noise Ratio (PSNR), 
and Mean Absolute Error (MAE). Perceptual quality is 
evaluated using Learned Perceptual Image Patch Similarity 
(LPIPS). All metrics are computed as 
averages over the full 0--3 hour prediction horizon unless 
otherwise specified.

\begin{table*}[t]
\small
\centering
\caption{Performance comparison of different models under skill scores on \textbf{Southeast China} datasets.
The best results are highlighted in bold, and the second-best results are underlined.}
\label{tab:se}
\resizebox{2\columnwidth}{!}{
\setlength{\tabcolsep}{2pt}
\renewcommand{\arraystretch}{1.2}
\begin{tabular}{@{}lcccccccccccccccc@{}}
\toprule
\multirow{2}{*}{\textbf{Models}} & \multirow{2}{*}{\textbf{Loss}} 
& \multicolumn{4}{c}{\textbf{$\uparrow$ CSI}} 
& \multicolumn{4}{c}{\textbf{$\uparrow$ ETS}}
& \multicolumn{4}{c}{\textbf{$\downarrow$ FAR}} 
& \multirow{2}{*}{\textbf{$\uparrow$ SSIM}} 
& \multirow{2}{*}{\textbf{$\uparrow$ PSNR}}
& \multirow{2}{*}{\textbf{$\downarrow$ MAE}}\\  
\cmidrule(lr){3-6} \cmidrule(lr){7-10} \cmidrule(lr){11-14}
& & $\gamma \ge 10$& $\gamma \ge 15$ & $\gamma \ge 20$ & $\gamma \ge 30$ 
& $\gamma \ge 10$& $\gamma \ge 15$ & $\gamma \ge 20$ & $\gamma \ge 30$
& $\gamma \ge 10$& $\gamma \ge 15$ & $\gamma \ge 20$ & $\gamma \ge 30$\\
\midrule
\midrule
OptFlow
&-
&0.548 & 0.513 & 0.458 & 0.342 &0.468 & 0.450 & 0.418 & 0.327
&0.303 & 0.329 & 0.372 & 0.482 &0.692 &19.374 &0.040 \\

Sprog
&- 
&\underline{0.561} & \underline{0.528} & \underline{0.471} & \underline{0.368} 
&0.486 & \underline{0.467} & 0.433 & \underline{0.354}
&0.266 & 0.307 & 0.352 & 0.446 &0.367 &12.160 &0.270 \\

TrajGRU
&MSE 
&0.557 & 0.518 & 0.470 & 0.339 
&\underline{0.491} & 0.466 & \underline{0.438} & 0.328   
&\underline{0.177} & \underline{0.192} & \underline{0.227} & \underline{0.270} 
&\underline{0.734} &\underline{20.843} &\underline{0.034} \\

AA-TransUnet
&MSE 
&0.549 & 0.515 & 0.466 & 0.349 
&0.483 & 0.463 & 0.434 & 0.339
&0.182 & 0.200 & 0.230 & 0.289 
&0.718 &20.725 &\underline{0.034} \\

Earthformer
&MSE 
&0.475 & 0.428 & 0.363 & 0.236 
&0.406 & 0.375 & 0.331 & 0.226 
&0.222 & 0.245 & 0.295 & 0.385 
&0.630 &19.182 &0.040 \\

MambaUnet
&MSE 
&0.509 & 0.461 & 0.412 & 0.284 
&0.442 & 0.409 & 0.380 & 0.274
&0.190 & 0.213 & 0.256 & 0.310 
&0.679 &20.021 &0.037 \\
\midrule

Diffcast
&-
&0.459 & 0.371 & 0.307 & 0.245
&0.401 & 0.327 & 0.266 & 0.212
&0.393 & 0.525 & 0.611 & 0.689
&0.387 &19.079 &2.148 \\
\midrule

Ours
&Spectral Loss
&\textbf{0.619} & \textbf{0.590} & \textbf{0.536} & \textbf{0.418} 
&\textbf{0.558} & \textbf{0.541} & \textbf{0.505} & \textbf{0.407}
&\textbf{0.149} & \textbf{0.169} & \textbf{0.205} & \textbf{0.262} 
&\textbf{0.788} &\textbf{21.696} &\textbf{0.030}\\
\bottomrule
\end{tabular}}
\end{table*}
\begin{table*}[t]
\small
\centering
\caption{Performance comparison of different models under skill scores on \textbf{Xinjiang} datasets.
The best results are highlighted in bold, and the second-best results are underlined.
}
\resizebox{2\columnwidth}{!}{  
\label{tab:xj}
\setlength{\tabcolsep}{2pt} 
\renewcommand{\arraystretch}{1.2} 
\begin{tabular}{@{}lccccccccccccc@{}}
\toprule
\multirow{2}{*}{\textbf{Models}} & \multirow{2}{*}{\textbf{Loss}} & \multicolumn{3}{c}{\textbf{$\uparrow$ CSI}} & \multicolumn{3}{c}{\textbf{$\uparrow$ ETS}}& \multicolumn{3}{c}{\textbf{$\downarrow$ FAR}} & \multicolumn{3}{c}{\textbf{$\uparrow$ POD}}\\  \cmidrule(lr){3-5} \cmidrule(lr){6-8} \cmidrule(lr){9-11}\cmidrule(lr){12-14}
& & $\gamma \ge 10$ & $\gamma \ge 15$ & $\gamma \ge 20$ 
& $\gamma \ge 10$ & $\gamma \ge 15$ & $\gamma \ge 20$ 
& $\gamma \ge 10$ & $\gamma \ge 15$ & $\gamma \ge 20$ 
& $\gamma \ge 10$ & $\gamma \ge 15$ & $\gamma \ge 20$ \\
\midrule
\midrule

OptFlow
&-
&  0.403& 0.343&  \underline{0.275} & 0.381&  0.328&  0.266
&  0.442& 0.505& 0.584 &\underline{0.591} &\underline{0.529}  &\underline{0.448} \\

Sprog
&-
&  \underline{0.453}	& \underline{0.414} &\textbf{0.357}  & \underline{0.434}& \underline{0.399} &\textbf{0.350}  				  				
& 0.364	& 0.422 &0.482  & \textbf{0.611}	& \textbf{0.593} &\textbf{0.537}  \\

TrajGRU
&MSE 
&  0.370& 0.328& 0.262 &0.355& 0.318& 0.256
&  0.225& 0.274& 0.352 & 0.415&0.375 &0.305 \\

AA-TransUnet
&MSE 
&  0.371& 0.328& 0.255 &0.356 &0.318 &0.249 
&  0.223& \underline{0.265}&0.347& 0.415& 0.372 & 0.295 \\
  
Earthformer
&MSE 
& 0.369 & 0.327 &0.195 &0.354& 0.316& 0.191
&0.240& 0.290& \textbf{0.325}  &0.417&0.377&0.216\\

MambaUnet
&MSE 
&0.329& 0.292& 0.209 &0.315& 0.282& 0.204
&\textbf{0.200}&  \textbf{0.259}&  0.394 & 0.358 &0.325 &0.242 \\
\cline{1-14}
\midrule

Ours
&Spectral Loss
& \textbf{0.464}&\textbf{0.428} & \textbf{0.357}& \textbf{0.449}& \textbf{0.417}& \textbf{0.350}
&\underline{0.220}& 0.276& \underline{0.342}  &0.537& 0.512 &0.438   \\
\bottomrule
\vspace{-0.3cm} 
\end{tabular}}
\end{table*}

\begin{table}[htbp]
  \centering
  \caption{Model parameters comparison on \textbf{Southeast China} datasets.}
  \resizebox{\columnwidth}{!}{  
  \begin{tabular}{@{}lcccc@{}}
    \toprule
    \textbf{Model}     & \textbf{Params} (M) & \textbf{Inference(ms)} $\downarrow$ & \textbf{CSI}($\gamma \geq$20)$\uparrow$& \textbf{CSI}($\gamma \geq$30)$\uparrow$\\
    \midrule
    TrajGRU   &11.92   &3965   &0.470   &0.339 \\
    AA‑TransUnet   &39.98   &448   &0.466   &0.349 \\
    Earthformer    &18.05   &233   &0.363   &0.236 \\
    MambaUnet    &8.97   &613   &0.412   &0.284 \\
    Diffcast       &50.45   &6345   &0.307   &0.245 \\
    \hline
    Ours  &20.36   &320   &0.536   &0.418 \\
    \bottomrule
  \end{tabular}}
  \vspace{-0.3cm} 
  \label{tab:params}
\end{table}

\noindent\textbf{Baseline Comparisons.} We compare MambaRain against representative methods including traditional optical flow approaches (OptFlow) and variational methods (Sprog), ConvRNN-based models (TrajGRU), CNN-Transformer hybrid architectures (AA-TransUnet), meteorology-oriented models (Earthformer), recent Mamba-based networks (MambaUnet~\cite{liu2024swin}), and diffusion-based Diffcast~\cite{yu2024diffcast}.

\subsection{Comparisons With State-of-the-Art Methods}
\begin{figure*}[ht]
\centering  
\includegraphics[width=1\textwidth]{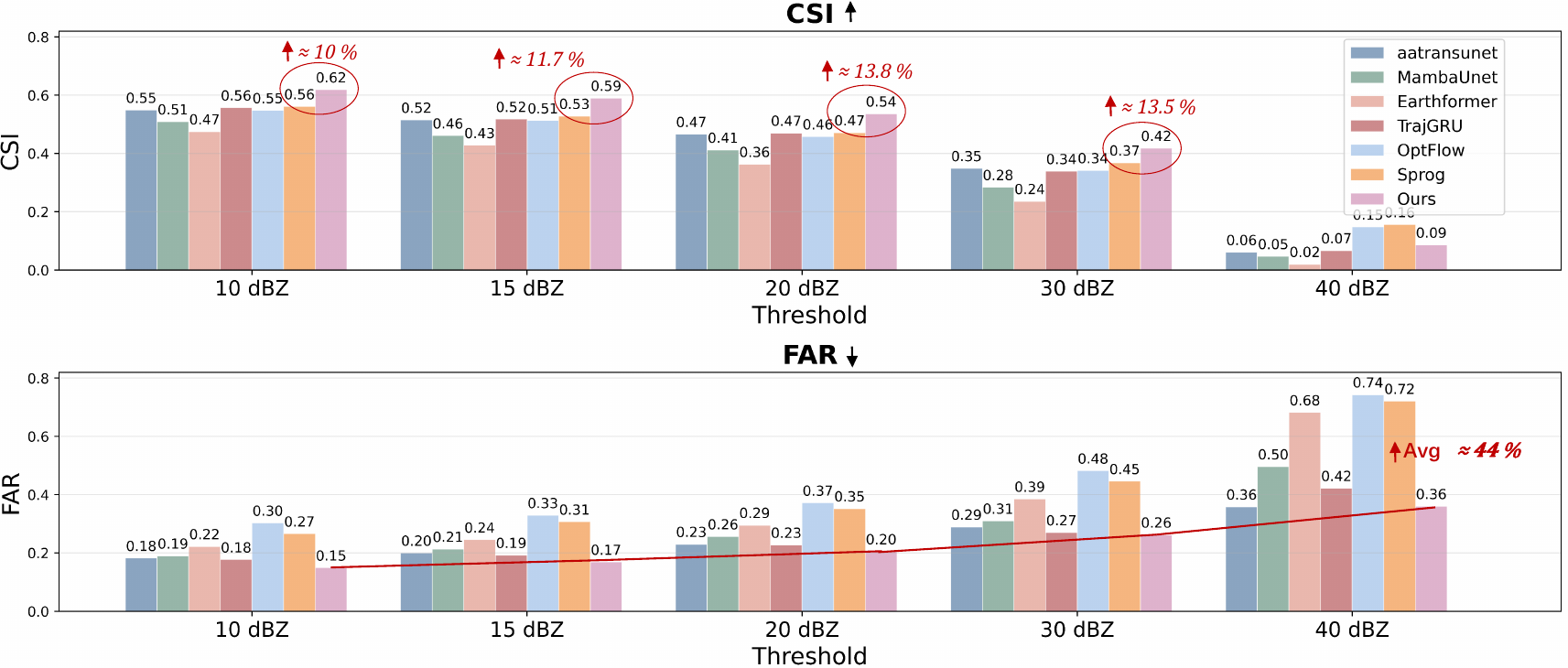}  
\caption{Threshold-dependent performance comparison of CSI 
($\uparrow$) and FAR ($\downarrow$) on the Southeast China dataset. 
MambaRain achieves consistent CSI improvements of 10--13.8\% over 
the best baseline across all thresholds, while maintaining the 
lowest FAR on average ($\approx$44\% reduction).}
\label{fig:zb}
\vspace{-2mm}
\end{figure*}
\begin{figure*}[ht]
\centering  
\includegraphics[width=1\textwidth]{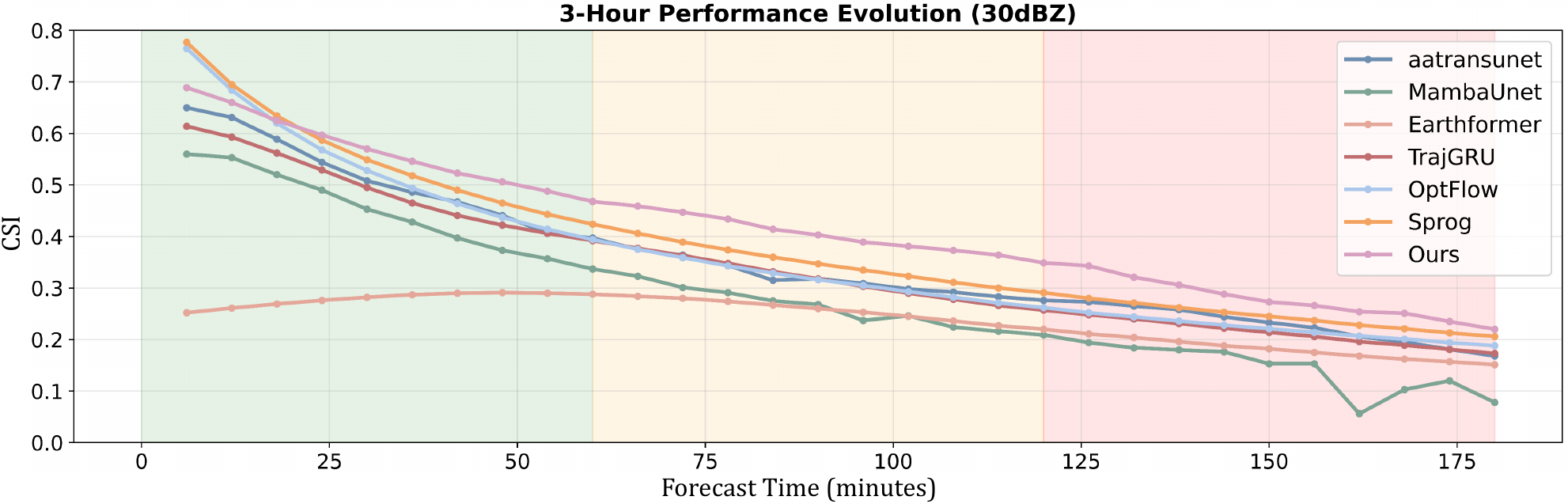}
\caption{Temporal evolution of CSI at the 30\,dBZ threshold over 
the 0--3 hour forecasting horizon on the Southeast China dataset. 
The shaded regions denote short-range (0--50\,min), medium-range 
(50--120\,min), and extended-range (120--180\,min) forecasting 
windows, respectively.}
\label{fig:30dBZ}
\vspace{-4mm}
\end{figure*}
\begin{figure*}[ht]
\centering  
\includegraphics[width=0.75\textwidth]{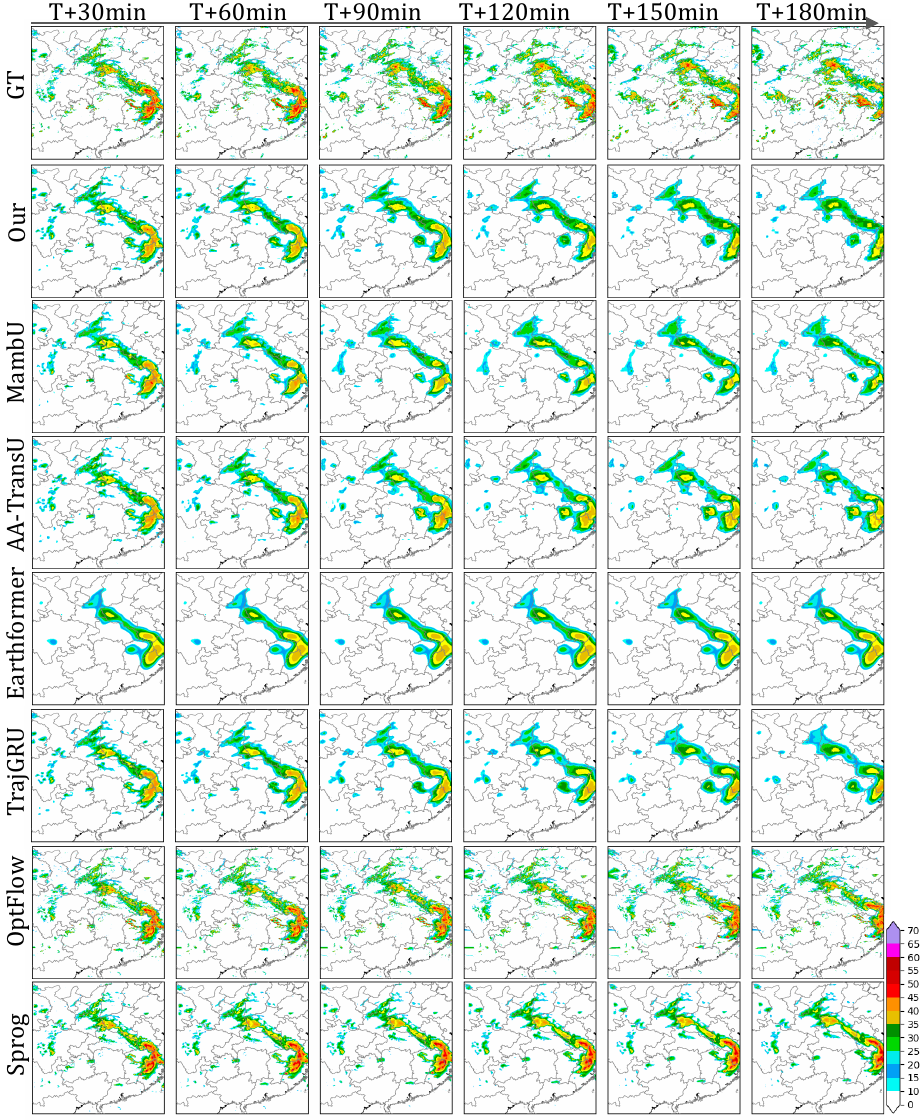}  
\caption{Visualization comparison of different models results at T+30 to T+180\,min on Southeast China dataset. Rows correspond to ground truth and seven competing methods; 
columns represent increasing forecast lead times.} 
\label{fig:vis}
\vspace{-2mm}
\end{figure*} 
\begin{figure*}[htb]
\centering  
\includegraphics[width=0.98\textwidth]{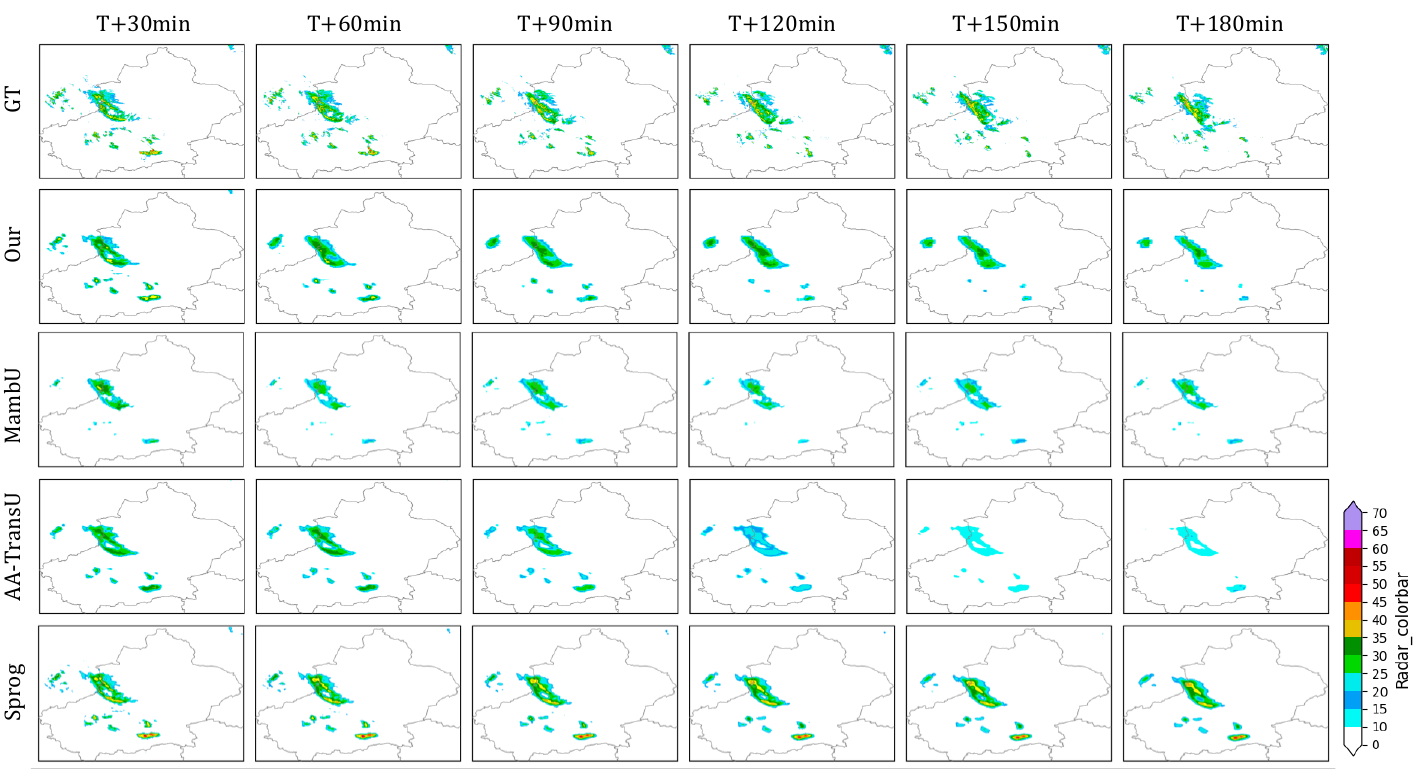}  
\caption{Visualization comparison of different models results at T+30 to T+180\,min on Xinjiang Province dataset. Rows correspond to ground truth and seven competing methods; 
columns represent increasing forecast lead times.} 
\label{fig:vis_xj}
\vspace{-6mm}
\end{figure*} 

\noindent\textbf{Quantitative Results.} 
Table~\ref{tab:se} presents performance comparisons on the Southeast 
China dataset. MambaRain consistently achieves superior performance 
across all meteorological and image quality metrics. Specifically, our 
method achieves CSI improvements of 11.7\%, 14.0\%, and 13.6\% at 
15\,dBZ, 20\,dBZ, and 30\,dBZ thresholds compared to the 
best-performing baselines, demonstrating substantial gains across the 
full intensity spectrum. The FAR reductions are particularly notable, 
averaging approximately 44\% relative to Sprog, which indicates that 
MambaRain produces significantly fewer false alarms while maintaining 
high detection rates. In terms of image quality, MambaRain attains 
the highest SSIM (0.788) and PSNR (21.696\,dB) with the lowest MAE 
(0.030), confirming that our framework preserves both structural 
fidelity and pixel-level accuracy. The comparison with the generative 
baseline DiffCast is especially revealing: MambaRain achieves 74.6\% 
higher CSI at $\gamma \geq 20$\,dBZ while reducing MAE by over 98\% 
and delivering approximately 20$\times$ faster inference. This 
demonstrates that our deterministic approach with spectral loss can 
surpass generative methods without the computational burden of 
iterative denoising processes.

The Xinjiang dataset presents a more challenging evaluation scenario, 
as precipitation events in this semi-arid mountainous region are 
characterized by overall weaker echo intensities and sparser 
occurrence compared to Southeast China. Despite this, as shown in 
Table~\ref{tab:xj}, MambaRain achieves state-of-the-art performance 
across all CSI and ETS thresholds, outperforming all competing methods 
including the strong traditional baseline Sprog. An interesting 
pattern emerges when examining the FAR-POD trade-off: deep learning 
baselines achieve lower FAR than traditional methods but at the cost 
of substantially reduced POD, indicating a systematic tendency toward 
underprediction. MambaRain strikes a better balance between these 
competing objectives, maintaining competitive FAR while achieving the 
highest POD among all deep learning methods. This balanced performance 
is particularly valuable in operational settings where both missed 
detections and false alarms carry significant costs. We attribute this 
advantage to the terrain-aware DEM encoding, which effectively anchors 
precipitation predictions to the region's complex orographic landscape 
and helps the model distinguish genuine orographic enhancement from 
spurious patterns.

\noindent\textbf{Computational Efficiency Analysis.} 
Table~\ref{tab:params} compares model parameters and inference 
latency across all methods on the Southeast China dataset. MambaRain 
achieves an inference time of \textbf{320\,ms} per sample with 
20.36M parameters, representing a favorable trade-off between 
accuracy and efficiency. Compared to DiffCast (50.45M parameters, 
6345\,ms inference), MambaRain is approximately \textbf{20$\times$} 
faster while achieving substantially higher CSI scores (0.536 vs. 
0.307 at $\gamma \geq 20$\,dBZ, a \textbf{74.6\%} improvement). 
Although Earthformer achieves the fastest inference (233\,ms) with 
fewer parameters (18.05M), its CSI scores are significantly lower 
(0.363 at $\gamma \geq 20$\,dBZ), indicating that its computational 
savings come at a considerable accuracy cost. TrajGRU, despite 
having only 11.92M parameters, requires 3965\,ms due to its 
sequential recurrent processing. MambaRain's 320\,ms latency 
comfortably meets the real-time operational requirement for 
6-minute update cycles, making it practically deployable in 
meteorological forecasting centers.

\noindent\textbf{Threshold-Dependent Analysis.} 
Fig.~\ref{fig:zb} illustrates CSI and FAR 
performance across different precipitation intensity thresholds on 
the Southeast China dataset. MambaRain maintains consistent 
advantages in both metrics across the entire threshold range from 
10\,dBZ to 50\,dBZ. The most substantial improvements are observed 
at moderate precipitation intensities (20--30\,dBZ), which correspond 
to light-to-moderate rainfall events that occur most frequently and 
are critical for practical operational applications. At higher 
thresholds ($\geq$40\,dBZ), all methods exhibit performance 
degradation due to the inherent rarity and high spatial variability 
of heavy precipitation events, but MambaRain still maintains a 
noticeable margin in both CSI and FAR over competing approaches, 
demonstrating its robustness across the full intensity spectrum.

\noindent\textbf{Temporal Performance Evolution.} Fig.~\ref{fig:30dBZ} demonstrates our model's sustained performance advantage throughout the 0-3 hour forecasting horizon. Unlike baseline methods that exhibit rapid performance degradation beyond 90 minutes, MambaRain maintains relatively stable CSI scores, with the 30dBZ threshold analysis revealing superior long-term forecasting capabilities essential for extended nowcasting applications.

\noindent\textbf{Qualitative Analysis.} 
Figs.~\ref{fig:vis} and~\ref{fig:vis_xj} provide visual comparisons 
of precipitation forecasts across six time horizons (T+30\,min to 
T+180\,min) on the Southeast China and Xinjiang datasets, 
respectively. On the Southeast China dataset (Fig.~\ref{fig:vis}), 
traditional methods OptFlow and Sprog produce reasonable spatial 
structures at short lead times but exhibit severe smoothing artifacts 
and echo dissipation beyond T+90\,min, as their linear motion 
assumptions fundamentally fail to capture convective initiation and 
cell merging at extended horizons. Deep learning baselines 
(TrajGRU, AA-TransUnet, Earthformer, MambaUnet) yield sharper 
outputs at early time steps but suffer from progressive intensity 
underestimation driven by the MSE regression-to-the-mean effect, 
with predicted echo fields fading noticeably by T+120\,min. In 
contrast, MambaRain maintains more realistic precipitation structures 
with better-retained echo intensities and sharper spatial boundaries 
throughout the full 3-hour window, directly benefiting from the 
spectral loss that penalizes high-frequency discrepancies in the 
Fourier domain. On the Xinjiang dataset (Fig.~\ref{fig:vis_xj}), 
where overall echo intensities are weaker and precipitation patterns 
are strongly modulated by complex orography, MambaUnet and 
AA-TransUnet still exhibit noticeable intensity underestimation at 
longer lead times, while Sprog remains relatively competitive at 
early horizons due to its simple extrapolation mechanism. MambaRain 
produces spatially coherent forecasts that better preserve 
orographic precipitation structures along the mountain ranges, 
demonstrating the practical value of terrain-aware DEM encoding 
in topographically complex regions.

\begin{table}[h!]
\centering
\caption{Ablation study on the Xinjiang and Southeast China dataset.
The best results are highlighted in bold, and the second-best results are underlined.
}
\resizebox{\columnwidth}{!}{  
\begin{tabular}{ccccccc}
\hline
\textbf{Block} & \textbf{Data} &\textbf{Loss} &\multicolumn{2}{c}{\textbf{Xinjiang}}&\multicolumn{2}{c}{\textbf{Southeast China}} \\ 
\cmidrule(lr){4-5} \cmidrule(lr){6-7}
MFormer & DEM & Spectral loss& Avg.CSI $\uparrow$ & Avg.FAR $\downarrow$ & Avg.CSI $\uparrow$ & Avg.FAR $\downarrow$ \\ \hline
\hline
$\checkmark$ & $\checkmark$ & $\times$& 0.263 & 0.383 & 0.488 & 0.210 \\
$\checkmark$ & $\times$ & $\checkmark$& 0.288 & 0.373 & 0.497 &  0.213\\
$\times$ & $\checkmark$ & $\checkmark$ & \underline{0.301} & \underline{0.326} &  \underline{0.507}& \underline{0.208} \\
\hline
\hline
$\checkmark$ & $\checkmark$ & $\checkmark$& \textbf{0.321} & \textbf{0.314}& \textbf{0.540} & \textbf{0.196} \\ \hline
\end{tabular}
\vspace{-0.9cm} 
\label{tab:ab}}
\end{table}
\noindent\textbf{Ablation Studies.} 
Table~\ref{tab:ab} presents ablation results on both datasets 
to validate each proposed component. The full MambaRain 
configuration consistently achieves the best performance in terms 
of both CSI and FAR. Removing the MFormer module induces the 
most severe degradation across both datasets, confirming that the 
hybrid Mamba-Attention design is the primary contributor to 
spatiotemporal modeling capability. Replacing the spectral loss 
with standard MSE leads to a notable CSI drop and increased FAR, 
validating that frequency-domain supervision is essential for 
preserving high-frequency precipitation structures at extended 
lead times. Excluding DEM encoding produces a more pronounced 
performance penalty on the Xinjiang dataset than on Southeast 
China, consistent with the expectation that terrain-aware 
geographic priors are especially critical in orographically 
complex regions. These results confirm that the three components 
are complementary and collectively necessary for achieving 
state-of-the-art nowcasting performance.

\section{Conclusion}
This paper presents MambaRain, a multi-scale precipitation 
nowcasting framework for extended 0--3 hour forecasting from 
ground-based radar remote sensing observations. The proposed 
MFormer block synergistically combines Mamba's linear-complexity 
sequential temporal modeling with self-attention's parallel 
spatial reasoning, addressing the complementary limitations of 
each mechanism when applied independently to spatiotemporal 
radar data. Terrain-aware DEM encoding is incorporated to 
mitigate systematic orographic biases in topographically complex 
regions, and a Fourier-domain spectral loss is introduced to 
preserve fine-scale precipitation structures against the 
regression-to-the-mean blurring effect inherent in 
MSE-optimized deterministic models. Extensive experiments on 
the geographically diverse Xinjiang and Southeast China SWAN 
datasets demonstrate that MambaRain consistently outperforms 
state-of-the-art baselines in both CSI and FAR across the full 
intensity threshold range, while meeting sub-second operational 
latency requirements. Ablation studies confirm the complementary 
contribution of each proposed component to the overall 
performance gains.

\noindent\textbf{Limitations and Future Work.}
Despite the promising results, the current framework is 
deterministic and does not quantify forecast uncertainty, 
and its generalization across heterogeneous radar networks 
remains to be verified. Future work will focus on further 
refinement of the proposed architecture, including more 
fine-grained spatiotemporal representation and improved 
handling of extreme precipitation at high intensity thresholds.

{
    \small
    \bibliographystyle{IEEEtran}
    \bibliography{main}
}
 
\end{document}